\pdfoutput=1
\documentclass[letterpaper,10pt,conference]{ieeeconf}
\IEEEoverridecommandlockouts
\usepackage[T1]{fontenc}
\usepackage[utf8]{inputenc}
\usepackage{amsmath,amssymb,bm}
\usepackage{graphicx}
\usepackage{cite}
\usepackage{array,booktabs,makecell}
\usepackage[dvipsnames]{xcolor}
\usepackage{caption}
\usepackage[hidelinks]{hyperref}
\newif\ifanonymous
\anonymousfalse 
\hypersetup{pdfsubject={},pdfkeywords={}}
\newcommand{\stdval}[1]{\,{\color{gray}\tiny$\pm$#1}}
\title{\LARGE\bf
Beyond Point-Attached Semantics: Stable Object-Centric Semantic Fields for Robust Manipulation
}
\ifanonymous
  \author{Anonymous Authors}
  \hypersetup{pdfauthor={}}
  \newcommand{\projectpage}{\url{https://anonymous.4open.science/w/beyond_point_attached_semantics/}}
\else
  \author{Zheng SUN$^{1}$, Lerong ZHANG$^{1}$, Zhihao LI$^{1}$, Zhuo LI$^{1}$, Quentin ROUXEL$^{1}$, and Fei CHEN$^{1,\dagger}$%
    \thanks{$^{1}$All authors are with The Chinese University of Hong Kong, Hong Kong SAR, China.}%
    \thanks{$^{\dagger}$Corresponding author: Fei Chen (e-mail: \texttt{f.chen@ieee.org}).}%
  }
  \hypersetup{pdfauthor={Zheng Sun, Zhihao Li, Zhuo Li, Quentin Rouxel, Fei Chen}}
  \newcommand{\projectpage}{\url{https://zainzh.github.io/beyond-point-attached-semantics/}}
\fi

\begin{document}
\bstctlcite{icra:compact-authors}
\maketitle
\thispagestyle{empty}
\pagestyle{empty}

\begin{abstract}
Robotic manipulation often requires identifying functional parts, such as a mug handle or a hammer head. However, features attached to observed 3D points can vary with viewpoint and sensor noise, giving a policy inconsistent representations of the same part. We propose an object-centric semantic field to provide more consistent part-aware features for manipulation. We use the observed object cloud to build a continuous field, then read features from this field at 3D locations independently resampled from the cloud. Each feature uses the sampled object support as context, rather than directly reusing an individual point descriptor. Part classification distinguishes functional regions, cross-instance alignment brings corresponding part features together, and perturbation consistency encourages similar features under observation changes. The queried coordinates and features form semantic point clouds that are supplied to a DP3-based policy. We evaluate the approach on four RoboTwin simulation tasks and four real-world bimanual tasks, achieving average success rates of 69.3\% and 67.5\%, respectively. These improve on Utonia Point-wise by 7.0 and 32.5 percentage points, respectively, with real-world tests on held-out objects. A point-wise control with matched part supervision scores 63.5\% in simulation, compared with our 69.3\%. These results highlight the value of stable, object-conditioned semantic fields for manipulation across object instances and varying observations. Project page: \projectpage.
\end{abstract}

\section{Introduction}

Robotic manipulation requires knowing not only where an object is, but which part to use. For example, striking a target with a hammer requires grasping its handle and making contact with its head. We refer to regions with distinct roles in an action as functional parts. Although their shapes and appearances vary across object instances, corresponding parts often serve similar roles. A policy therefore needs a representation that consistently identifies corresponding functional parts across object instances and observations, helping it learn which parts to grasp or use.

Such a representation is not directly available to the policy; it must be derived from sensor observations. Point clouds provide explicit 3D geometry for manipulation~\cite{ze20243d,ke20243d}, but their samples vary with viewpoint, depth noise, and sampling procedure. Lifting image features into 3D~\cite{chen2025g3flow,fan2026any3d,wang2024gendp} or using pretrained point-cloud encoders~\cite{zhang2026utonia} can add part-related information, yet does not necessarily remove this observation dependence. Even when an object and its function remain unchanged, the same part can have different features across observations. These action-irrelevant changes can make it harder for the policy to learn consistent associations between parts and actions from limited demonstrations.
\begin{figure}[t]
    \centering
    \includegraphics[width=\linewidth]{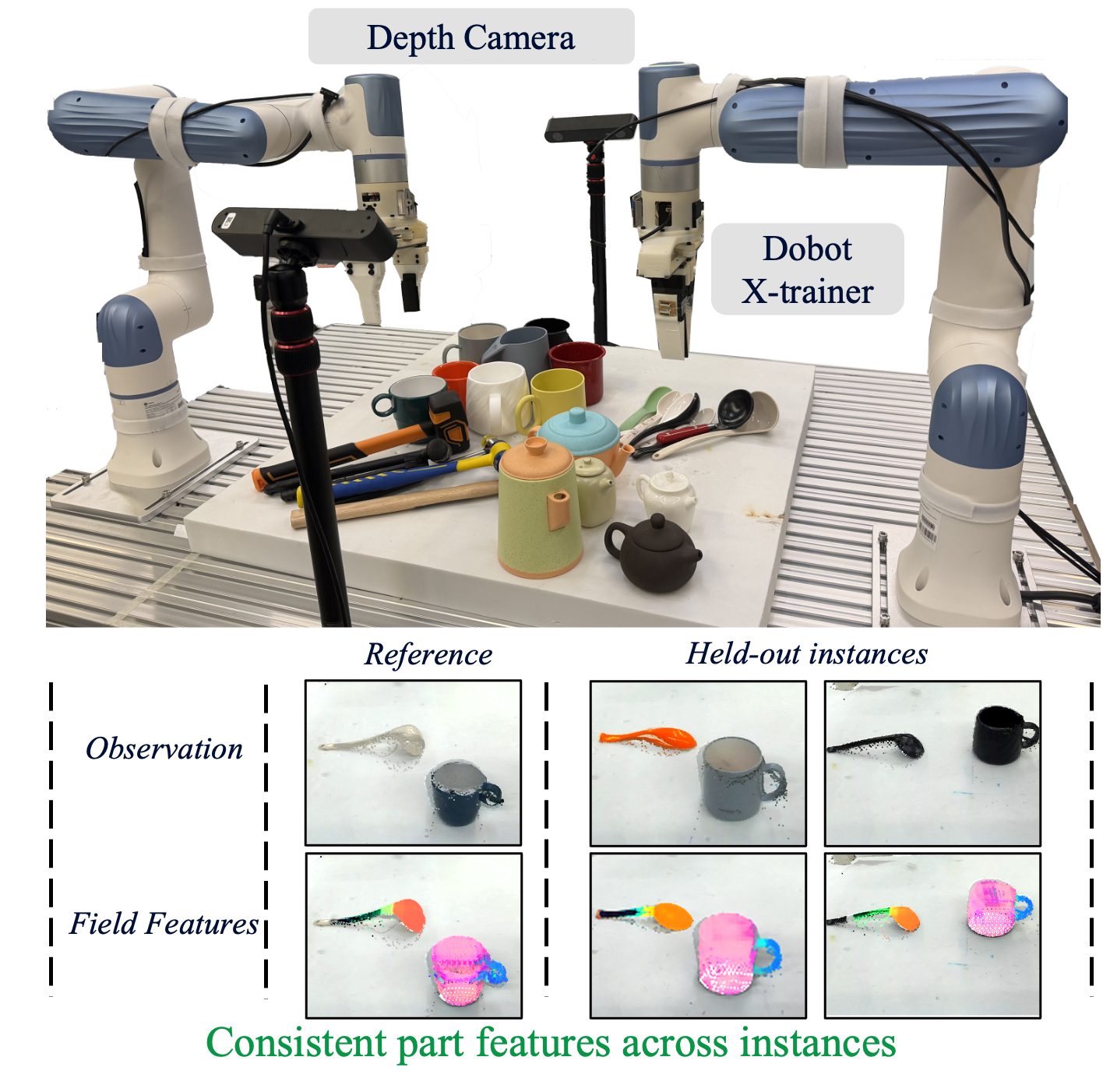}
    \caption{Real-world platform and cross-instance part features. Top: the DOBOT X-Trainer with two RGB-D cameras. Bottom: observed point clouds and our field features for a reference scene and held-out object instances. Corresponding mug and spoon parts exhibit similar feature colors across instances, while different parts remain distinguishable.}
    \label{fig:real_world_experiment_setup}
\end{figure}

\begin{figure*}[t]
    \centering
    \includegraphics[width=1\linewidth]{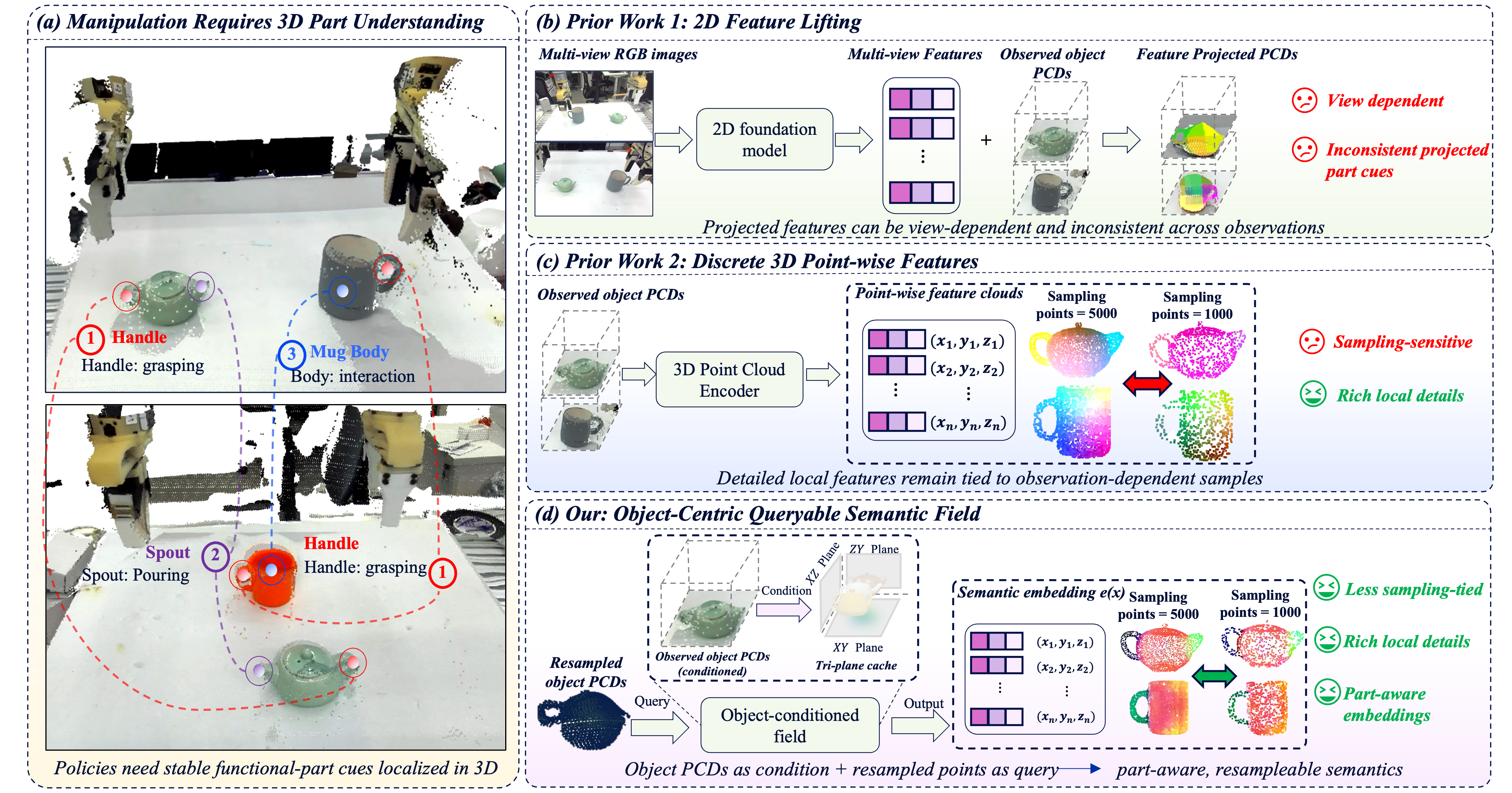}
    \caption{Summary of our object-centric semantic field. Point-attached baselines associate features with individual observed samples. Our field uses the observed object cloud to infer part-aware features at resampled 3D locations. The resulting coordinates and features form semantic point clouds for manipulation.}
    \label{fig:teaser}
\end{figure*}

To reduce this burden, we seek a representation that distinguishes functional parts, associates corresponding parts across object instances, and is less sensitive to observation changes. For example, a mug handle should have similar features across views and mugs, while remaining distinct from the mug body. We propose an object-centric continuous semantic field to learn these properties (Fig.~\ref{fig:teaser}). The field predicts part features at 3D locations independently resampled from the observed object cloud. Each feature is inferred using the sampled object support as context, rather than reusing an individual input-point descriptor.


We learn these properties from part-annotated 3D models in PartNext~\cite{wang2026partnext}, training one field per category. A frozen Utonia encoder~\cite{zhang2026utonia} extracts geometric features from the observed cloud, providing a geometric prior for the field. Part labels teach the field to distinguish regions such as handles and bodies. We also encourage corresponding parts on different objects to have similar features, and train with perturbed observations to discourage feature changes caused by input variation.

The learned field improves the object representation supplied to a DP3-based policy without changing its action representation or learning objective. We freeze the field after training and combine its queried features with their 3D coordinates to form a semantic point cloud at each policy step. The policy receives this cloud as additional object information and learns from manipulation demonstrations. Part annotations are used to train the field on 3D models, not to label the demonstrations.

We evaluate the approach on four RoboTwin simulation tasks~\cite{chen2025robotwin} and four real-world bimanual tasks, achieving average success rates of 69.3\% and 67.5\%, respectively. These improve on Utonia Point-wise by 7.0 and 32.5 percentage points, respectively, with real-world tests on held-out objects. Under matched part supervision, the field also exceeds a point-wise control by 5.8 points in simulation. Feature visualizations examine part consistency, and loss ablations assess the training objectives.
\begin{itemize}
    \item An object-centric continuous semantic field that infers features at resampled 3D locations using object-level support context, aiming to provide consistent part information to a manipulation policy.
    \item A field-learning formulation that combines part supervision, cross-instance alignment, and perturbation consistency to preserve part distinctions across varying instances and observations.
    \item Simulation and real-world evaluations demonstrating improved manipulation success with the field, supported by representation analysis, a supervised point-wise control, and ablations of the learning objectives.
\end{itemize}

\section{Related Work}

\subsection{3D Visuomotor Policy Learning}

Imitation learning acquires manipulation skills from visual and proprioceptive demonstrations through behavioral cloning~\cite{florence2022implicit,li2025manidp} or diffusion-based action generation~\cite{chi2025diffusion,rouxel2024flow}. Recent policies make spatial structure explicit using point clouds~\cite{ze20243d,ke20243d}, voxel grids~\cite{shridhar2023perceiver}, or geometrically related multi-view features~\cite{goyal2023rvt}. Point clouds are particularly attractive because they preserve metric coordinates through a compact policy interface. However, raw point clouds do not identify functional parts, and RGB-D samples vary with viewpoint, occlusion, segmentation, and resampling. Twin-DP3 addresses viewpoint changes by reconstructing digital-twin point clouds from object models, pose tracking, and robot geometry~\cite{liu2026twindp3}. We instead learn part features from the current observation while retaining the policy objective.

\subsection{Semantic 3D Representations for Manipulation}
\looseness=-1

Semantic 3D representations supplement geometry with functional cues. One family of methods lifts pretrained 2D descriptors such as DINOv2~\cite{oquab2023dinov2} into 3D, including Distilled Feature Fields~\cite{shen2023distilled}, G3Flow~\cite{chen2025g3flow}, and GenDP~\cite{wang2024gendp}. Another family learns affordances, action trajectories, or dense part-aware features directly from 3D data~\cite{mo2021where2act,wu2021vat,xu2026action}. Large-scale point-cloud encoders~\cite{xie2020pointcontrast,zhang2026utonia} and fine-grained part datasets such as PartNext~\cite{wang2026partnext} further support transferable geometric and part understanding. HeRO combines lifted DINOv2 and Stable Diffusion features with hierarchical global/local policy conditioning~\cite{xu2026hero}. Our method instead uses labeled 3D parts to train a category-level field and tests this representation under matched supervision.

\subsection{Queryable Object-Centric Fields}
Object-centric correspondences provide a structured basis for category-level manipulation. kPAM represents manipulation-relevant keypoint affordances~\cite{manuelli2019kpam}, TAX-Pose estimates task-specific cross-object transformations~\cite{pan2023tax}, and Neural Descriptor Fields provide continuous SE(3)-equivariant descriptors~\cite{simeonov2022neural}. Dense feature fields, including D3Fields~\cite{wang2023d}, Distilled Feature Fields~\cite{shen2023distilled}, GenDP~\cite{wang2024gendp}, and G3Flow~\cite{chen2025g3flow}, further demonstrate the utility of queryable 3D semantics for learned manipulation. Continuous querying is therefore not our contribution alone. We learn an observation-conditioned field with category-level part supervision, cross-instance alignment, and perturbation consistency. Unlike direct point-wise prediction, the field aggregates support features into a spatial cache and decodes embeddings at resampled queries. Our supervised point-wise control tests the value of this representation beyond the shared encoder and training labels.

\section{Methodology}
\label{sec:method}

\subsection{Problem Formulation}

\begin{figure*}[t]
    \centering
    \includegraphics[width=1\linewidth]{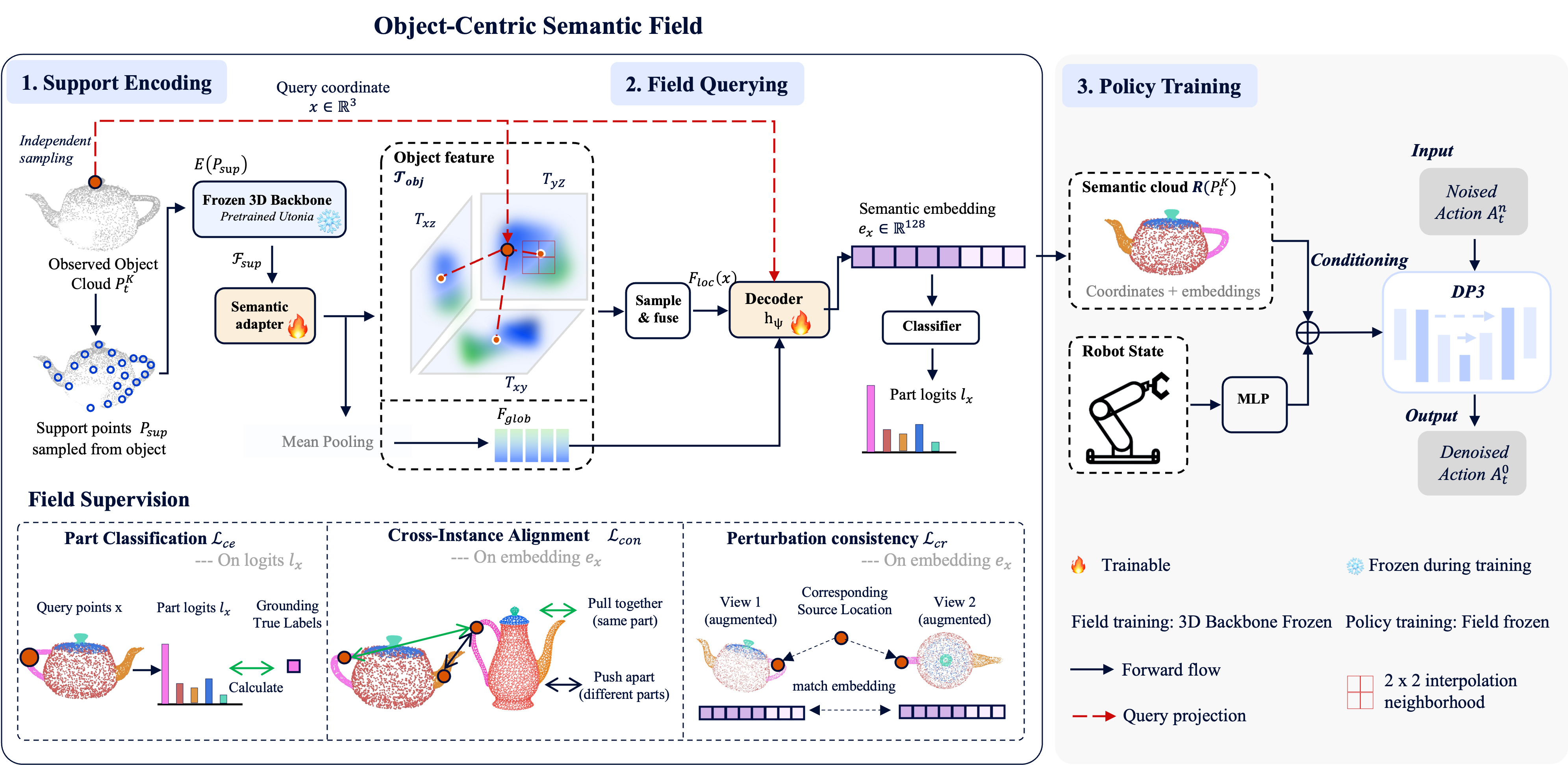}
    \caption{Overview of our object-centric continuous semantic field. Support points condition an object-specific tri-plane cache; 3D query locations read out part-aware embeddings and training-time part logits. The frozen field is then queried at resampled object locations to export semantic point clouds for downstream manipulation policies.}
    \label{fig:method}
\end{figure*}

Our goal is to learn an object representation with consistent part semantics for robust manipulation across instances within the same category. We aim to reduce observation-induced feature variation unrelated to the required action, helping the policy learn associations between parts and actions from demonstrations. A representation function $R$ converts each observed object cloud $\mathcal{P}_t^k$ into a semantic point cloud containing 3D coordinates and part features. At time $t$, the policy combines these representations for $K$ task-relevant objects with the scene cloud $\mathcal{C}_t$ and robot proprioception $q_t$ to predict an action sequence $A_t$:
\begin{equation}
    A_t\sim\pi\!\left(\cdot\mid\mathcal{C}_t,q_t,
    \{R(\mathcal{P}_t^k)\}_{k=1}^{K}\right),
    \label{eq:policy_problem}
\end{equation}
The policy's action representation and diffusion learning objective remain unchanged. To implement $R$, we construct a continuous semantic field conditioned on the observed object cloud and query it at 3D locations to obtain part features. We train the field on part-annotated 3D object models, then freeze it and train the policy on manipulation demonstrations. Fig.~\ref{fig:method} summarizes the method.

\subsection{Semantic Field Construction}
\label{sec:field_construction}

To reduce reliance on descriptors attached to individual observed points, we first build a semantic field from a set of object points, then read features from the field. We call the points used to build it the support set $\mathcal{P}_{\mathrm{sup}}$. Given a query location $x$, the field predicts a part feature:
\begin{equation}
    e_x=f_{\theta}(x\mid\mathcal{P}_{\mathrm{sup}}).
    \label{eq:field}
\end{equation}
Here, $\theta$ denotes the trainable parameters used to compute the feature. While $f_{\theta}$ outputs one query feature, $R$ in Eq.~\eqref{eq:policy_problem} pairs features from multiple queries with their world-frame positions to form the object representation. At deployment, we reconstruct each object cloud from depth pixels selected by Grounded SAM masks in calibrated RGB-D views. Support and query points are independently sampled from the current object cloud $\mathcal{P}_t^k$; the support need not contain all observed points, and queries need not remain fixed across observations. Both sets share a support-derived transform: subtract the support centroid, divide by the maximum support radius, then translate the support's horizontal bounding-box center and minimum height to the origin. The field equations use these normalized coordinates.

We first extract support features using a frozen Utonia encoder $E$~\cite{zhang2026utonia}, which provides a pretrained geometric prior. A learnable semantic adapter $g_{\phi}$ adjusts these features for our part-learning objectives:
\begin{equation}
    \widetilde{\mathcal{F}}_{\mathrm{sup}}
    =g_{\phi}\!\left(E(\mathcal{P}_{\mathrm{sup}})\right).
    \label{eq:support_encoding}
\end{equation}
The adapter uses two 256-channel linear layers with LayerNorm and GELU. Utonia receives coordinates, colors, and normals, with voxel size 0.01 in the normalized frame.

The adapted features still lie at discrete support points. To read them at resampled locations, we project them onto the $xy$, $xz$, and $yz$ planes and distribute them to neighboring grid cells with bilinear weights. We also average the adapted support features to obtain a global vector $F_{\mathrm{glob}}$. We collect the three feature grids and this global vector into an object feature cache, denoted by $\mathcal{T}_{\mathrm{obj}}$:
\begin{equation}
    \mathcal{T}_{\mathrm{obj}}
    =\{T_{xy},T_{xz},T_{yz},F_{\mathrm{glob}}\}.
    \label{eq:triplane}
\end{equation}
Here, $T_{xy}$, $T_{xz}$, and $T_{yz}$ are 2D feature grids on the $xy$, $xz$, and $yz$ planes, respectively. Each grid has 64-by-64 cells and 256 channels. Projection and querying share support bounds expanded by 5\%; each cell stores a weighted average of nearby projected features. Queries share the cache for each observation; new support points trigger its reconstruction.

For a query $x$, we read from the three grids in $\mathcal{T}_{\mathrm{obj}}$ by projecting its coordinate onto each plane using the same support bounds as plane construction. We read features by bilinear interpolation, concatenate them, and apply a learnable fusion block to obtain the local feature $F_{\mathrm{loc}}(x)$. A linear layer with LayerNorm and GELU fuses the three sampled features into 256 dimensions. This local feature varies with the query, while the 256-D global average supplies shared object context. The decoder uses both to infer the part feature at the query coordinate:
\begin{equation}
\begin{aligned}
    e_x&=h_{\psi}\!\left(x,F_{\mathrm{loc}}(x),F_{\mathrm{glob}}\right),\\
    \ell_x&=W_{\mathrm{cls}}e_x+b_{\mathrm{cls}}.
\end{aligned}
\label{eq:query_decoder}
\end{equation}
The decoder uses two 256-channel layers with LayerNorm and GELU, followed by a linear projection to the 128-D L2-normalized part feature $e_x$. A linear head then maps it to class scores $\ell_x$ for part supervision. The policy uses the part features; the class scores are needed only during field training. The adapter, local fusion module, and decoder together have parameters $\theta$; the classification head has weights $W_{\mathrm{cls}}$ and bias $b_{\mathrm{cls}}$.

\subsection{Semantic Field Learning}
\label{sec:objectives}

We train the network in Sec.~\ref{sec:field_construction} using PartNext part labels~\cite{wang2026partnext}, with one field per object category. For each model, we sample support points and labeled queries, then generate two augmented views using random rotations and coordinate jitter. These are transformations of the sampled points, not separate camera observations, and both use the same network. Query index $j$ tracks the same source surface location, producing features $e_j^{(1)},e_j^{(2)}$ and class scores $\ell_j^{(1)},\ell_j^{(2)}$, with part label $y_j$.

Three losses supervise the query outputs to distinguish parts, align corresponding parts across instances, and reduce feature changes under input perturbations.

\textbf{Part classification.}
To distinguish labeled parts, $\mathcal{L}_{\mathrm{ce}}$ applies cross-entropy to the class scores $\ell_j^{(v)}$ and labels $y_j$, averaged over queries and both views:
\begin{equation}
    \mathcal{L}_{\mathrm{ce}}
    =\frac{1}{2N_q}\sum_{v=1}^{2}\sum_{j=1}^{N_q}
    \mathrm{CE}(\ell_j^{(v)},y_j),
    \label{eq:part_loss}
\end{equation}
where $N_q$ is the number of training queries per object. We average this per-object loss over the minibatch.

\textbf{Cross-instance alignment.}
We apply supervised contrastive learning to $e_x$: queries with the same part label are positives, including those from different object instances, while different parts are negatives. For one augmented view, we omit the view index and compute
\begin{equation}
\begin{aligned}
\mathcal{L}_{\mathrm{con}}^{\mathrm{view}}
&=-\frac{1}{|\mathcal{I}|}\sum_{i\in\mathcal{I}}\frac{1}{|\mathcal{P}^{+}(i)|}\\
&\quad\times\sum_{p\in\mathcal{P}^{+}(i)}\log
\frac{\exp(e_i^\top e_p/\tau)}{\sum_{a\in\mathcal{A}\setminus\{i\}}\exp(e_i^\top e_a/\tau)},
\end{aligned}
\label{eq:align_loss}
\end{equation}
where $\mathcal{A}$ indexes retained queries across the minibatch, $\mathcal{P}^{+}(i)$ contains other queries with the same label as $i$, $\mathcal{I}$ contains anchors with at least one positive, and $\tau$ is the temperature. We average the two view losses.

\textbf{Perturbation consistency.}
$\mathcal{L}_{\mathrm{cr}}$ also acts on embeddings, but compares the same source location across augmentations:
\begin{equation}
    \mathcal{L}_{\mathrm{cr}}
    =\frac{1}{N_q}\sum_{j=1}^{N_q}
    \left(1-\langle e_j^{(1)},e_j^{(2)}\rangle\right).
    \label{eq:stability_loss}
\end{equation}
Because the embeddings are L2-normalized, this loss penalizes their cosine dissimilarity, encouraging consistent part features under the applied rotations and coordinate jitter. We average it over objects in the minibatch.

We optimize the combined objective:
\begin{equation}
    \mathcal{L}=\lambda_{\mathrm{ce}}\mathcal{L}_{\mathrm{ce}}
    +\lambda_{\mathrm{con}}\mathcal{L}_{\mathrm{con}}
    +\lambda_{\mathrm{cr}}\mathcal{L}_{\mathrm{cr}}.
    \label{eq:total_loss}
\end{equation}
All three losses update the adapter, local fusion module, and decoder through the query outputs and differentiable grid operations. Only the classification loss also updates the linear classification head. Utonia remains frozen, and the feature grids are computed from the support rather than learned as independent parameters. After field training, we freeze the field for policy learning.

We sample 5,000 support points by surface area and 2,048 labeled queries per training object. Query sampling combines 75\% part-balanced and 25\% area-weighted samples. After normalization, each view uses an independent uniform SO(3) rotation and Gaussian coordinate jitter with standard deviation 0.005. The classification, contrastive, and consistency weights are 1.0, 0.2, and 0.1, respectively, with contrastive temperature 0.1. We retain at most 2,048 valid queries per view for contrastive learning. Training uses AdamW for 4,000 epochs, batch size 6, learning rate 0.001, weight decay 0.0001, and mixed precision.

\subsection{Policy Learning with Semantic Point Clouds}
\label{sec:method_policy}

We use DP3~\cite{ze20243d} as the manipulation policy backbone. Separate PointNets encode the scene cloud and each object's semantic cloud. Each semantic point contains its 3D position and part feature $e_x$; a point-wise MLP, max pooling, and projection produce an object-level feature. We concatenate the scene and object features with an MLP encoding of robot proprioception $q_t$, then flatten the features across the observation history into a global condition. This condition modulates the action-denoising network through FiLM to predict the action sequence $A_t$.

We train the policy to predict clean, normalized demonstration actions from their noisy versions, conditioned on the observations. The objective is mean squared error over action dimensions and sequence steps. Training updates the policy encoders and denoiser while keeping the semantic field frozen. At deployment, the policy executes a prefix of the predicted action sequence and replans from new observations. All controlled comparisons retain the same action representation and learning objective.

\section{Experiments}
\label{sec:experiments}

We evaluate three questions:
\begin{enumerate}
    \renewcommand{\labelenumi}{(\arabic{enumi})}
    \item \textbf{Representation Quality:} Do features distinguish parts and align corresponding regions across instances?
    \item \textbf{Manipulation Effectiveness:} Do they improve success in simulation and on held-out real objects?
    \item \textbf{Component Contributions:} What do field-based readout and each learning objective contribute?
\end{enumerate}

\subsection{Experimental Setup}
\begin{figure}[t]
    \centering
    \includegraphics[width=\linewidth]{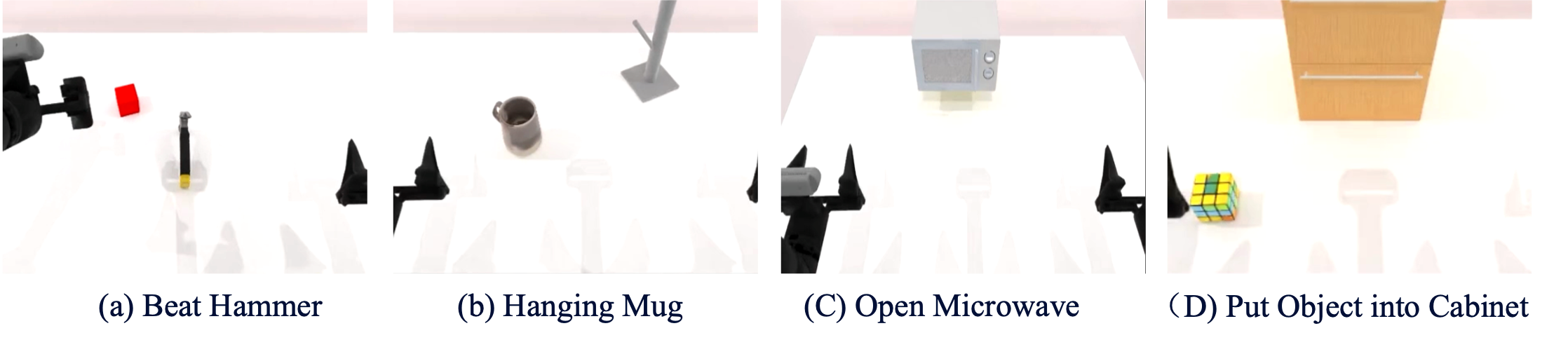}
    \caption{The four RoboTwin tasks: Beat Hammer, Hanging Mug, Open Microwave, and Put Object into Cabinet.}
    \label{fig:simulation_tasks}
\end{figure}

\begin{figure}[t]
    \centering
    \includegraphics[width=\linewidth]{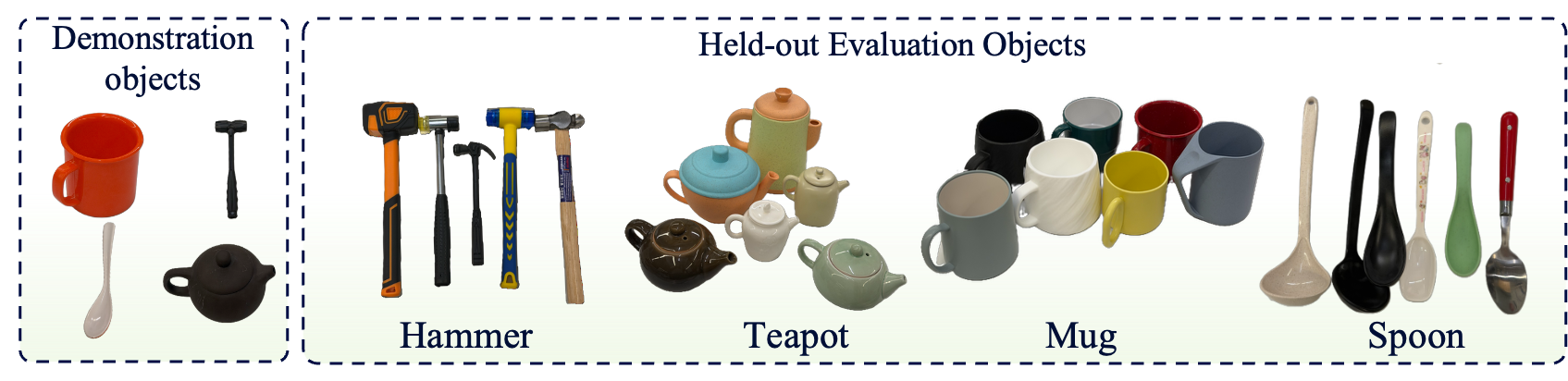}
    \caption{Objects used in real-world experiments. Left: objects used to collect policy-training demonstrations. Right: held-out hammer, teapot, mug, and spoon instances used for evaluation, with variations in geometry and appearance.}
    \label{fig:real_world_experiment_objects}
\end{figure}

\paragraph{Simulation setup.}
We evaluate four RoboTwin tasks~\cite{chen2025robotwin} spanning single-arm and dual-arm manipulation: \emph{Hanging Mug} hangs a mug on a rack through its handle; \emph{Beat Hammer} grasps the handle and strikes a target with the hammer head; \emph{Open Microwave} pulls the door handle; and \emph{Put Object into Cabinet} grasps the cabinet handle, opens the door, and places the object inside. These tasks require functional-part localization (Fig.~\ref{fig:simulation_tasks}). Simulation training and evaluation use the same task and object setup under the benchmark protocol, without an additional held-out-object split. All methods use the same 50 demonstrations and test episodes. Policies train for 3,000 epochs, with 100 test episodes per seed.

\paragraph{Real-world setup.}
We evaluate four bimanual tasks on a DOBOT X-Trainer with two ZED 2i RGB-D cameras (Fig.~\ref{fig:real_world_experiment_setup}). We collect 50 demonstrations per task using designated training objects. Policy evaluation uses only held-out objects, disjoint from those in the demonstrations (Fig.~\ref{fig:real_world_experiment_objects}). Success requires: (1) \emph{Grasp Mug}, grasping the handle and placing the mug stably on the base; (2) \emph{Beat Cube}, grasping the hammer handle and striking the cube with its head; (3) \emph{Stir Mug}, holding the mug handle with the left hand and spoon handle with the right, then stirring with the spoon head; (4) \emph{Pour Water}, grasping the teapot handle and pouring from its spout into the mug. Each method runs 20 trials per task. Task objects are randomly selected from the held-out set, with a 100-second limit per rollout.

Demonstrations use one training instance per category. The held-out sets contain five hammers, six teapots, six mugs, and six spoons. For Stir Mug and Pour Water, both objects are drawn from their respective held-out categories. Success rates use 20 trials per task, not per instance or pair; sampling need not cover instances equally.

\paragraph{Baselines.}
\textbf{DP3}~\cite{ze20243d} is the geometry-only baseline. \textbf{DINOv2 Lifting} projects frozen image descriptors~\cite{oquab2023dinov2} onto observed object points, inspired by G3Flow~\cite{chen2025g3flow} and D$^3$Fields~\cite{wang2023d}. \textbf{Utonia Point-wise} attaches frozen 3D descriptors~\cite{zhang2026utonia} directly to observed points. We also evaluate the official \textbf{G3Flow}~\cite{chen2025g3flow} pipeline in simulation, retaining its object-model-based semantic flow and FoundationPose tracking. DINOv2 Lifting is a controlled feature baseline, not a reproduction of G3Flow.

\paragraph{Implementation and evaluation details.}
Grounded SAM~\cite{ren2024groundedsam} provides object masks. Calibrated depth is back-projected and fused in the robot/world frame; DINOv2 Lifting, Utonia Point-wise, and ours share this reconstruction pipeline and DP3 encoding pathway. Separate PointNets encode the scene and semantic object clouds, with unchanged action representation and learning objective. The scene input contains 1,024 points; our field uses 1,024 supports and 256 queries per object.

Each object PointNet produces a 128-D vector; its input descriptor dimension depends on the baseline. The scene encoder preserves surrounding geometry.

We train one field per task-relevant category on part-annotated PartNext models, without scans or object-specific fine-tuning from any physical evaluation or demonstration object. Real-world policies learn from demonstrations with the training objects and are evaluated on held-out instances.

All methods use matched demonstration budgets and evaluation protocols; controlled baselines also share observation histories and training schedules. Policies train for 3,000 epochs on an RTX 4090 using AdamW, learning rate 0.0001, weight decay 0.000001, and cosine decay with 500 warm-up steps. Batch size is 128. From three observation steps, the policy predicts eight action steps and executes six before replanning, using 10 DDIM denoising steps. Evaluation uses the epoch-3,000 checkpoint.

Simulation reports mean success and standard deviation across five training seeds, with 100 test episodes per task and seed. For each ablation, we first average the four task success rates within each seed, then report the mean and standard deviation across these five averages.

\subsection{Simulation Experiments}
\label{sec:simulation_results}

\begin{table}[t]
\centering
\scriptsize
\setlength{\tabcolsep}{1.0pt}
\captionsetup{justification=centering,labelsep=newline,font=footnotesize}
\caption{\textsc{Comparative Evaluation on Four RoboTwin Tasks (Success Rates in \%).} Task entries report mean and standard deviation across five training seeds.}
\label{tab:simulation_results}
\begin{tabular*}{\columnwidth}{@{\extracolsep{\fill}}l|cccc|c@{}}
\toprule
\textbf{Method}
& \makecell{\textbf{\textit{Hang}}\\\textbf{\textit{Mug}}}
& \makecell{\textbf{\textit{Beat}}\\\textbf{\textit{Hammer}}}
& \makecell{\textbf{\textit{Open}}\\\textbf{\textit{Micro.}}}
& \makecell{\textbf{\textit{Put}}\\\textbf{\textit{Cabinet}}}
& \textbf{\textit{Average}} \\
\midrule
DP3~\cite{ze20243d} & 23.0\stdval{4.0} & 72.0\stdval{2.0
} & 62.0\stdval{3.0} & 72.0\stdval{2.0} & 57.3\\
DINOv2 Lifting & 23.0\stdval{7.0} & 78.0\stdval{5.0} & 58.0\stdval{3.0} & 68.0\stdval{4.0} & 56.8 \\
Utonia Point-wise & 31.0\stdval{4.0} & 77.0\stdval{2.0} & 65.0\stdval{1.0} & 76.0\stdval{1.0} & 62.3 \\
G3Flow~\cite{chen2025g3flow} & 25.0\stdval{5.0} & 78.0\stdval{5.0} & 64.0\stdval{2.0} & 78.0\stdval{3.0} & 61.3 \\
\midrule
\textbf{Ours} & \textbf{37.0}\stdval{3.0} & \textbf{84.0}\stdval{1.0} & \textbf{73.0}\stdval{1.0} & \textbf{83.0}\stdval{1.0} & \textbf{69.3} \\
\bottomrule
\end{tabular*}
\end{table}

Table~\ref{tab:simulation_results} shows the highest success on every task for our method, averaging 69.3\%: 12.0 points above DP3, 7.0 above Utonia Point-wise, and 8.0 above G3Flow. The gains span handle localization, tool use, and articulated interaction. Random resampling changes the input point sets during evaluation; success-rate standard deviations of 1--3 points indicate consistent policy performance across training seeds, but do not directly measure feature stability.

DINOv2 Lifting improves Beat Hammer but matches or underperforms DP3 elsewhere, averaging 56.8\%. Its larger standard deviations on several tasks indicate greater training-run variability, without isolating multi-view fusion as the cause. Utonia Point-wise improves the DP3 average by 5.0 points; our field adds another 7.0. The supervised point-wise ablation below separates the benefit of field representation from additional part supervision.

G3Flow averages 61.3\%, exceeding DINOv2 Lifting by 4.5 points but remaining below ours on every task. Its standard deviations are 2--5 points, versus our 1--3. We observed FoundationPose drift that displaced object-aligned features, a possible contributor to this variability. Our field instead reads features from each current observation without propagating tracked poses. This comparison evaluates complete pipelines, not the effect of tracking in isolation.

\subsection{Real-World Experiments}
\label{sec:real_results}

\begin{figure}[t]
    \centering
    \includegraphics[width=\linewidth]{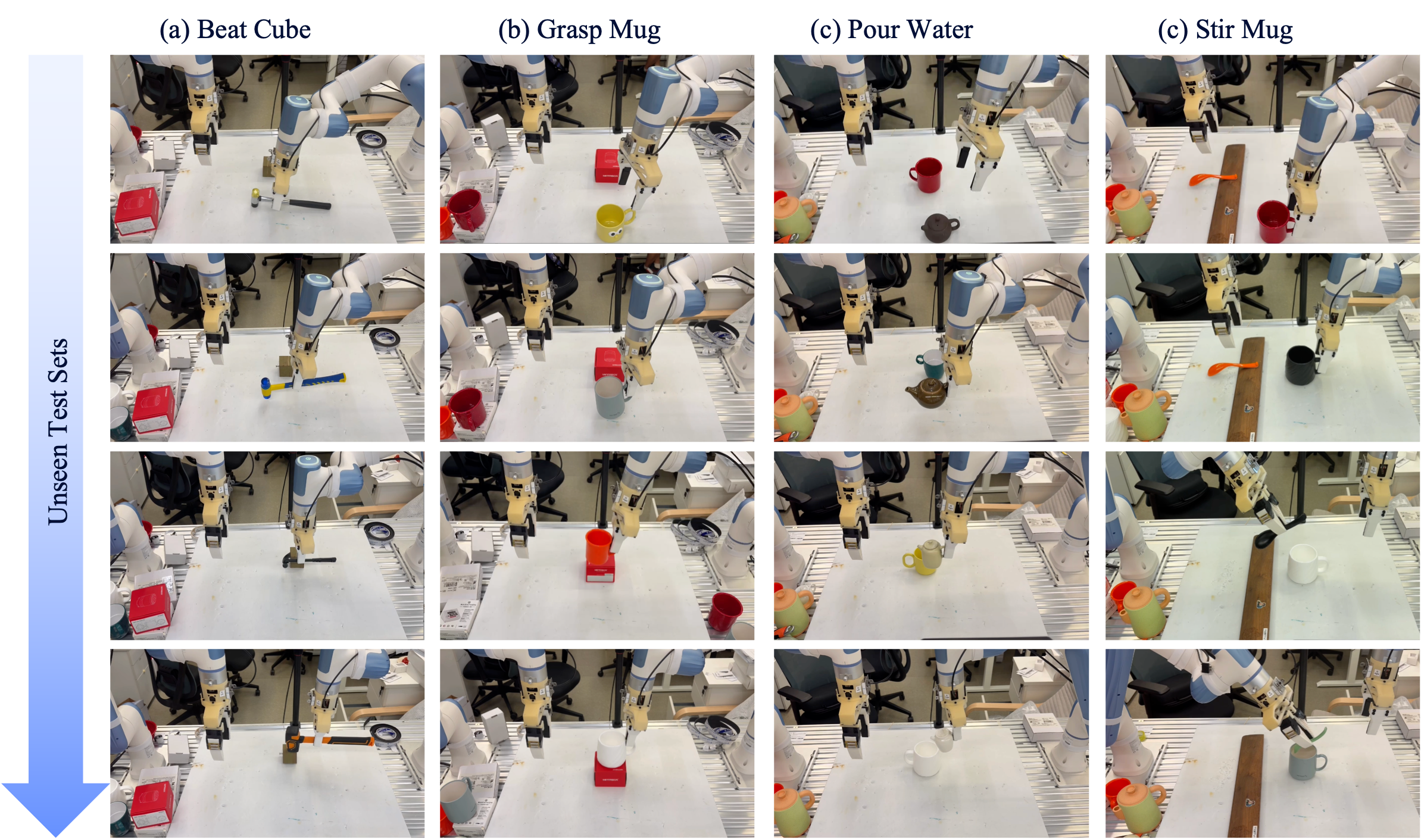}
    \caption{Real-world manipulation examples on held-out instances: Beat Cube, Grasp Mug, Pour Water, and Stir Mug.}
    \label{fig:real_world_tasks}
\end{figure}

\begin{table}[t]
\centering
\scriptsize
\setlength{\tabcolsep}{1.0pt}
\captionsetup{justification=centering,labelsep=newline,font=footnotesize}
\caption{\textsc{Comparative Evaluation on Four Real-World Tasks (Success Rates in \%).} Each method is evaluated for 20 trials per task.}
\label{tab:real_results}
\begin{tabular*}{\columnwidth}{@{\extracolsep{\fill}}l|cccc|c@{}}
\toprule
\textbf{Method}
& \makecell{\textbf{\textit{Grasp}}\\\textbf{\textit{Mug}}}
& \makecell{\textbf{\textit{Beat}}\\\textbf{\textit{Cube}}}
& \makecell{\textbf{\textit{Stir}}\\\textbf{\textit{Mug}}}
& \makecell{\textbf{\textit{Pour}}\\\textbf{\textit{Water}}}
& \textbf{\textit{Average}} \\
\midrule
DP3~\cite{ze20243d} & 35.0 & 40.0 & 15.0 & 20.0 & 27.5 \\
DINOv2 Lifting & 35.0 & 45.0 & 25.0 & 25.0 & 32.5 \\
Utonia Point-wise & 35.0 & 45.0 & 35.0 & 25.0 & 35.0 \\
\midrule
\textbf{Ours} & \textbf{85.0} & \textbf{85.0} & \textbf{50.0} & \textbf{50.0} & \textbf{67.5} \\
\bottomrule
\end{tabular*}
\end{table}

Fig.~\ref{fig:real_world_tasks} shows held-out-object executions. Our method leads all four tasks (Table~\ref{tab:real_results}), averaging 67.5\%, compared with 35.0\% for Utonia Point-wise, 32.5\% for DINOv2 Lifting, and 27.5\% for DP3. On Grasp Mug and Beat Cube, our method succeeds in 17/20 trials each; Utonia Point-wise succeeds in 7/20 and 9/20, respectively.

Stir Mug and Pour Water each use two held-out objects together. Our method succeeds in 10/20 trials on each, compared with 7/20 and 5/20 for Utonia Point-wise. The gains extend to complementary roles, such as holding a mug while stirring with a spoon.

The margin over Utonia Point-wise is larger in the real world than in simulation (32.5 versus 7.0 points). Sensing noise and partial visibility may destabilize point-attached features, motivating our object-conditioned readout. However, the tasks and object distributions also differ, so this gap cannot be attributed solely to sensor noise. The following analysis examines qualitative representation patterns.

\subsection{Representation Analysis}
\label{sec:representation_stability}

We examine part discrimination within instances and part consistency across instances. For each method and category, one PCA basis is fitted jointly to all displayed embeddings and applied without instance-specific normalization. Colors can thus be compared across instances within that group, but not across methods or categories. Similar colors suggest similar projected features; these visualizations do not quantify distances in the full embedding space.

\paragraph{Simulation representation.}

\begin{figure}[t]
    \centering
    \includegraphics[width=\linewidth]{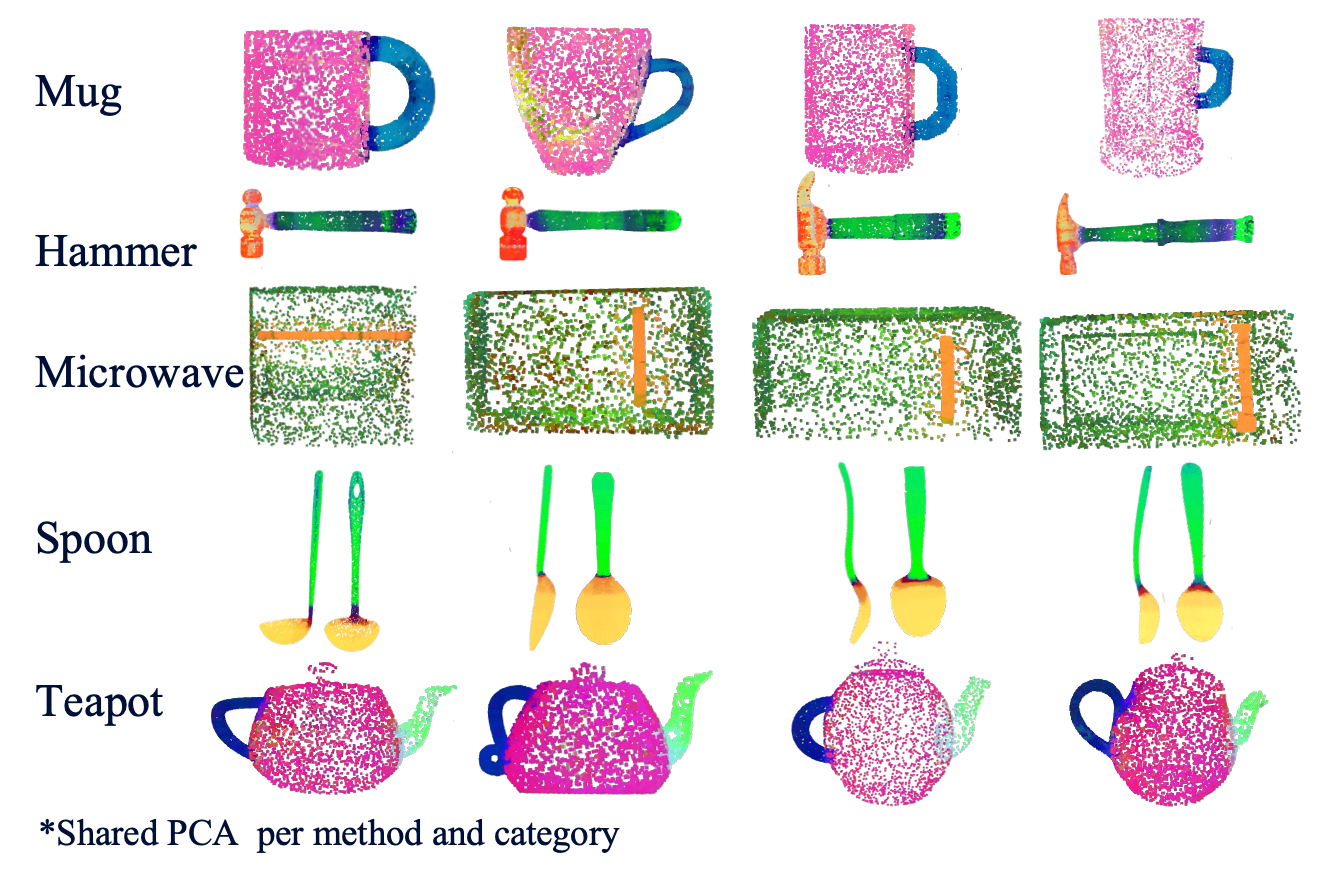}
    \caption{Part-aware field embeddings queried on simulation instances from five object categories. Within each category, all instances share one PCA projection. Corresponding functional parts exhibit consistent colors across instances, while distinct parts remain clearly separable.}
    \label{fig:simulation_feature}
\end{figure}

Fig.~\ref{fig:simulation_feature} shows five categories with substantial shape variation. Corresponding regions retain similar colors, while mug handles remain distinct from bodies and hammer heads from handles. These patterns suggest consistent part semantics across instances.

Consistency is part-specific, not a uniform feature over the entire object: corresponding handles should have similar embeddings while remaining distinct from bodies. Each query retains its 3D coordinates, preserving the spatial information needed to select grasp and contact locations.

\begin{figure}[t]
    \centering
    \includegraphics[width=\linewidth]{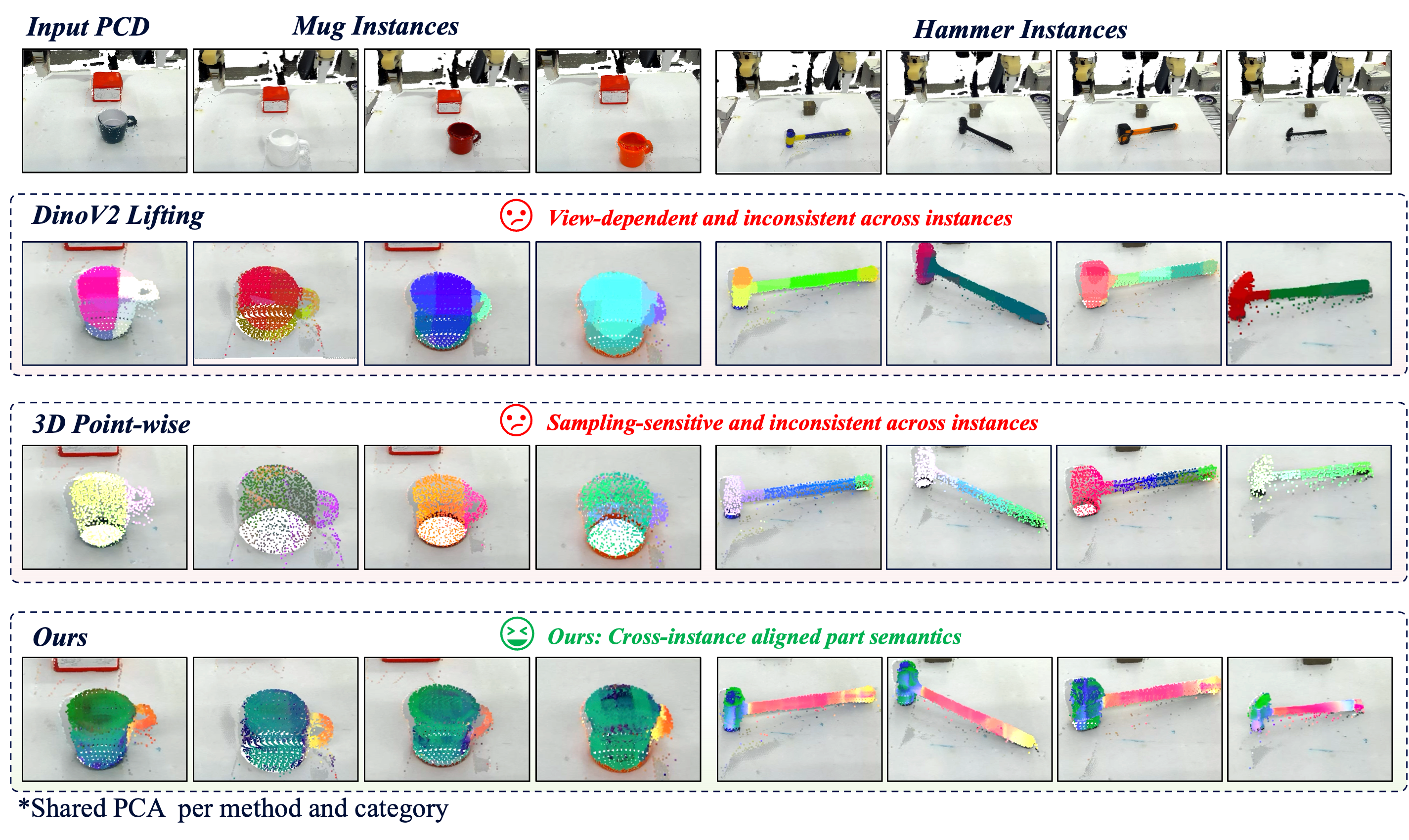}
    \caption{Features on real mug and hammer observations. Corresponding parts show more coherent color patterns with our field. Each method-category group uses a shared PCA basis; colors are not comparable between methods.}
    \label{fig:real_world_feature}
\end{figure}

\paragraph{Real-world representation.}
Fig.~\ref{fig:real_world_feature} compares real mug and hammer observations. DINOv2 Lifting often separates parts within one object, but corresponding regions show different colors across instances and views. Utonia Point-wise preserves recognizable geometry yet also exhibits color variation. Our field shows coherent handle/body and head/handle patterns across the displayed instances. These observations support part-aware cross-instance organization, but cannot isolate viewpoint, sampling, or noise effects.

\subsection{Ablation Study}
\label{sec:ablation}

Part-supervised Point-wise uses the same frozen Utonia encoder, part-training data, and three objectives, but predicts embeddings from observed point features without tri-plane construction or field querying. It samples observed points with their embeddings; the full model queries independently resampled locations. Both retain the same PointNet pathway, embedding dimension, DP3 backbone, action objective, and demonstrations. Unlike frozen Utonia Point-wise, this control learns from part labels. Other variants retain the field and remove one loss at a time.

\begin{table}[t]
\centering
\scriptsize
\setlength{\tabcolsep}{1.0pt}
\captionsetup{justification=centering,labelsep=newline,font=footnotesize}
\caption{\textsc{Ablation Study of Representation Design and Field-Learning Objectives.} Mean and standard deviation of four-task average success (\%) across five training seeds.}
\label{tab:ablation}
\begin{tabular*}{\columnwidth}{@{\extracolsep{\fill}}l|c|ccc|c@{}}
\toprule
\textbf{Variant}
& \makecell{\textbf{Field}\\\textbf{Readout}}
& $\boldsymbol{\mathcal{L}_{\mathrm{ce}}}$
& $\boldsymbol{\mathcal{L}_{\mathrm{con}}}$
& $\boldsymbol{\mathcal{L}_{\mathrm{cr}}}$
& \makecell{\textbf{\textit{Average}}\\\textbf{\textit{(\%)}}} \\
\midrule
\makecell[l]{Part-supervised\\Point-wise} &  & $\checkmark$ & $\checkmark$ & $\checkmark$ & 63.5\stdval{4.2} \\
\midrule
Without $\mathcal{L}_{\mathrm{ce}}$ & $\checkmark$ &  & $\checkmark$ & $\checkmark$ & 61.3\stdval{1.7} \\
Without $\mathcal{L}_{\mathrm{con}}$ & $\checkmark$ & $\checkmark$ &  & $\checkmark$ & 64.7\stdval{2.1} \\
Without $\mathcal{L}_{\mathrm{cr}}$ & $\checkmark$ & $\checkmark$ & $\checkmark$ &  & 68.9\stdval{1.8} \\
\midrule
\textbf{Full model} & $\checkmark$ & $\checkmark$ & $\checkmark$ & $\checkmark$ & \textbf{69.3}\stdval{1.0} \\
\bottomrule
\end{tabular*}
\end{table}

Table~\ref{tab:ablation} reports 69.3\% for the full field and 63.5\% for Part-supervised Point-wise, a 5.8-point gain under matched part supervision. This supports field-based feature construction and readout over directly predicting features at observed points. It evaluates the field representation as a whole, not independent query sampling alone.

The full model also has a smaller across-seed standard deviation than the supervised point-wise control (1.0 versus 4.2 points). Its higher mean is thus accompanied by less training-run variation under shared supervision. This comparison does not separate the effects of spatial aggregation, decoding, and query selection.

Removing part classification reduces success to 61.3\% (8.0 points), and removing cross-instance alignment yields 64.7\% (4.6 points). These are the largest observed losses among the training objectives. Without perturbation consistency, success is 68.9\%, only 0.4 points below the full model. This gap is smaller than the respective across-seed standard deviations of 1.8 and 1.0 points, so the summary statistics alone do not establish a reliable success-rate benefit from this objective. The ablations measure policy success, not feature stability.

\section{Discussion and Limitations}

Under the same part supervision, the full model outperforms the point-wise representation, supporting the additional value of field-based feature construction and readout. Random resampling exposes the policy to changing point sets during evaluation. In simulation, the full model's success-rate standard deviations are 1--3 percentage points across training seeds. This indicates consistent policy performance, but does not isolate feature robustness to specific disturbances. Controlled tests of depth noise, occlusion, and calibration errors would assess robustness beyond resampling.

\looseness=-1
Our method currently targets manipulation within known object categories, with a separate semantic field trained for each category using part-annotated 3D models. Extending it to new categories therefore requires suitable part labels and field training, but no additional part annotations on manipulation demonstrations. Sharing a field across categories and reducing the need for part supervision are directions for broader applicability. Although the field can generate features from partial observations, its predictions still depend on the available object information; severe occlusion or segmentation errors may affect part identification. Incorporating observation uncertainty could help the policy recognize and handle insufficient object information.

\section{Conclusion}
We presented an object-centric continuous semantic field that separates the points used to observe an object from the locations where policy-facing semantics are read. Conditioned on a partial object point cloud, the field produces part-aware embeddings at resampled query locations and exports them as a semantic point cloud to a DP3-based policy without changing its action objective. Part classification and cross-instance alignment organize features by functional region, while perturbation consistency encourages similar features under changes to the support observation. Across four RoboTwin tasks and four real-world bimanual tasks, the resulting representation improves manipulation success over raw point clouds, lifted 2D features, and point-wise 3D features, with larger margins over these baselines in the real-world evaluation. Ablations show that part classification and cross-instance alignment provide the largest observed benefits among the learning objectives. Under the same part supervision, the full field also outperforms a point-wise representation, supporting the value of field-based semantic readout. These results support object-conditioned semantic fields as useful representations for manipulation.

\bibliographystyle{IEEEtran}
\bibliography{bibliography_control,reference}

@IEEEtranBSTCTL{icra:compact-authors,
  CTLuse_forced_etal = "yes",
  CTLmax_names_forced_etal = "3",
  CTLnames_show_etal = "1"
}

@inproceedings{simeonov2022neural,
  title={Neural descriptor fields: Se (3)-equivariant object representations for manipulation},
  author={Simeonov, Anthony and Du, Yilun and Tagliasacchi, Andrea and Tenenbaum, Joshua B and Rodriguez, Alberto and Agrawal, Pulkit and Sitzmann, Vincent},
  booktitle={2022 International Conference on Robotics and Automation (ICRA)},
  pages={6394--6400},
  year={2022},
  organization={IEEE}
}

@inproceedings{manuelli2019kpam,
  title={kpam: Keypoint affordances for category-level robotic manipulation},
  author={Manuelli, Lucas and Gao, Wei and Florence, Peter and Tedrake, Russ},
  booktitle={The International Symposium of Robotics Research},
  pages={132--157},
  year={2019},
  organization={Springer}
}

@inproceedings{pan2023tax,
  title={Tax-pose: Task-specific cross-pose estimation for robot manipulation},
  author={Pan, Chuer and Okorn, Brian and Zhang, Harry and Eisner, Ben and Held, David},
  booktitle={Conference on Robot Learning},
  pages={1783--1792},
  year={2023},
  organization={PMLR}
}

@article{zhang2026utonia,
  title={Utonia: Toward One Encoder for All Point Clouds},
  author={Zhang, Yujia and Wu, Xiaoyang and Yang, Yunhan and Fan, Xianzhe and Li, Han and Zhang, Yuechen and Huang, Zehao and Wang, Naiyan and Zhao, Hengshuang},
  journal={arXiv preprint arXiv:2603.03283},
  year={2026}
}

@article{ze20243d,
  title={3d diffusion policy: Generalizable visuomotor policy learning via simple 3d representations},
  author={Ze, Yanjie and Zhang, Gu and Zhang, Kangning and Hu, Chenyuan and Wang, Muhan and Xu, Huazhe},
  journal={arXiv preprint arXiv:2403.03954},
  year={2024}
}

@article{ke20243d,
  title={3d diffuser actor: Policy diffusion with 3d scene representations},
  author={Ke, Tsung-Wei and Gkanatsios, Nikolaos and Fragkiadaki, Katerina},
  journal={arXiv preprint arXiv:2402.10885},
  year={2024}
}

@inproceedings{goyal2023rvt,
  title={Rvt: Robotic view transformer for 3d object manipulation},
  author={Goyal, Ankit and Xu, Jie and Guo, Yijie and Blukis, Valts and Chao, Yu-Wei and Fox, Dieter},
  booktitle={Conference on Robot Learning},
  pages={694--710},
  year={2023},
  organization={PMLR}
}

@inproceedings{chen2025g3flow,
  title={G3flow: Generative 3d semantic flow for pose-aware and generalizable object manipulation},
  author={Chen, Tianxing and Mu, Yao and Liang, Zhixuan and Chen, Zanxin and Peng, Shijia and Chen, Qiangyu and Xu, Mingkun and Hu, Ruizhen and Zhang, Hongyuan and Li, Xuelong and others},
  booktitle={Proceedings of the Computer Vision and Pattern Recognition Conference},
  pages={1735--1744},
  year={2025}
}

@article{fan2026any3d,
  title={Any3D-VLA: Enhancing VLA Robustness via Diverse Point Clouds},
  author={Fan, Xianzhe and Deng, Shengliang and Wu, Xiaoyang and Lu, Yuxiang and Li, Zhuoling and Yan, Mi and Zhang, Yujia and Zhang, Zhizheng and Wang, He and Zhao, Hengshuang},
  journal={arXiv preprint arXiv:2602.00807},
  year={2026}
}

@inproceedings{wang2024gendp,
  title={GenDP: 3D Semantic Fields for Category-Level Generalizable Diffusion Policy.},
  author={Wang, Yixuan and Yin, Guang and Huang, Binghao and Kelestemur, Tarik and Wang, Jiuguang and Li, Yunzhu},
  booktitle={CoRL},
  pages={4866--4878},
  year={2024}
}

@article{xu2026action,
  title={Action-Geometry Prediction with 3D Geometric Prior for Bimanual Manipulation},
  author={Xu, Chongyang and Li, Haipeng and Cheng, Shen and Hu, Jingyu and Fan, Haoqiang and Feng, Ziliang and Liu, Shuaicheng},
  journal={arXiv preprint arXiv:2602.23814},
  year={2026}
}

@inproceedings{xie2020pointcontrast,
  title={Pointcontrast: Unsupervised pre-training for 3d point cloud understanding},
  author={Xie, Saining and Gu, Jiatao and Guo, Demi and Qi, Charles R and Guibas, Leonidas and Litany, Or},
  booktitle={European conference on computer vision},
  pages={574--591},
  year={2020},
  organization={Springer}
}

@article{chi2025diffusion,
  title={Diffusion policy: Visuomotor policy learning via action diffusion},
  author={Chi, Cheng and Xu, Zhenjia and Feng, Siyuan and Cousineau, Eric and Du, Yilun and Burchfiel, Benjamin and Tedrake, Russ and Song, Shuran},
  journal={The International Journal of Robotics Research},
  volume={44},
  number={10-11},
  pages={1684--1704},
  year={2025},
  publisher={Sage Publications Sage UK: London, England}
}

@inproceedings{florence2022implicit,
  title={Implicit behavioral cloning},
  author={Florence, Pete and Lynch, Corey and Zeng, Andy and Ramirez, Oscar A and Wahid, Ayzaan and Downs, Laura and Wong, Adrian and Lee, Johnny and Mordatch, Igor and Tompson, Jonathan},
  booktitle={Conference on robot learning},
  pages={158--168},
  year={2022},
  organization={PMLR}
}

@inproceedings{shridhar2023perceiver,
  title={Perceiver-actor: A multi-task transformer for robotic manipulation},
  author={Shridhar, Mohit and Manuelli, Lucas and Fox, Dieter},
  booktitle={Conference on Robot Learning},
  pages={785--799},
  year={2023},
  organization={PMLR}
}

@article{oquab2023dinov2,
  title={Dinov2: Learning robust visual features without supervision},
  author={Oquab, Maxime and Darcet, Timoth{\'e}e and Moutakanni, Th{\'e}o and Vo, Huy and Szafraniec, Marc and Khalidov, Vasil and Fernandez, Pierre and Haziza, Daniel and Massa, Francisco and El-Nouby, Alaaeldin and others},
  journal={arXiv preprint arXiv:2304.07193},
  year={2023}
}

@article{shen2023distilled,
  title={Distilled feature fields enable few-shot language-guided manipulation},
  author={Shen, William and Yang, Ge and Yu, Alan and Wong, Jansen and Kaelbling, Leslie Pack and Isola, Phillip},
  journal={arXiv preprint arXiv:2308.07931},
  year={2023}
}

@inproceedings{wang2023d,
  title={{D3Fields}: Dynamic 3D Descriptor Fields for Zero-Shot Generalizable Robotic Manipulation},
  author={Wang, Yixuan and Zhang, Mingtong and Li, Zhuoran and Driggs-Campbell, Katherine Rose and Wu, Jiajun and Fei-Fei, Li and Li, Yunzhu},
  booktitle={ICRA 2024 Workshop on 3D Visual Representations for Robot Manipulation},
  year={2023}
}

@inproceedings{mo2021where2act,
  title={Where2act: From pixels to actions for articulated 3d objects},
  author={Mo, Kaichun and Guibas, Leonidas J and Mukadam, Mustafa and Gupta, Abhinav and Tulsiani, Shubham},
  booktitle={Proceedings of the IEEE/CVF International Conference on Computer Vision},
  pages={6813--6823},
  year={2021}
}

@article{wu2021vat,
  title={Vat-mart: Learning visual action trajectory proposals for manipulating 3d articulated objects},
  author={Wu, Ruihai and Zhao, Yan and Mo, Kaichun and Guo, Zizheng and Wang, Yian and Wu, Tianhao and Fan, Qingnan and Chen, Xuelin and Guibas, Leonidas and Dong, Hao},
  journal={arXiv preprint arXiv:2106.14440},
  year={2021}
}

@article{chen2025robotwin,
  title={Robotwin 2.0: A scalable data generator and benchmark with strong domain randomization for robust bimanual robotic manipulation},
  author={Chen, Tianxing and Chen, Zanxin and Chen, Baijun and Cai, Zijian and Liu, Yibin and Li, Zixuan and Liang, Qiwei and Lin, Xianliang and Ge, Yiheng and Gu, Zhenyu and others},
  journal={arXiv preprint arXiv:2506.18088},
  year={2025}
}

@article{wang2026partnext,
  title={Partnext: A next-generation dataset for fine-grained and hierarchical 3d part understanding},
  author={Wang, Penghao and He, Yiyang and Lv, Xin and Zhou, Yukai and Xu, Lan and Yu, Jingyi and Gu, Jiayuan},
  journal={Advances in Neural Information Processing Systems},
  volume={38},
  year={2026}
}

@article{ren2024groundedsam,
  title={Grounded {SAM}: Assembling Open-World Models for Diverse Visual Tasks},
  author={Ren, Tianhe and Liu, Shilong and Zeng, Ailing and Lin, Jing and Li, Kunkun and Cao, He and Chen, Jiayu and Huang, Xinyang and Chen, Yukang and Yan, Feng and Zeng, Zhaoyang and Zhang, Hao and Li, Feng and Yang, Jie and Li, Hongyang and Jiang, Qing and Zhang, Lei},
  journal={arXiv preprint arXiv:2401.14159},
  year={2024}
}

@inproceedings{rouxel2024flow,
  title={Flow matching imitation learning for multi-support manipulation},
  author={Rouxel, Quentin and Ferrari, Andrea and Ivaldi, Serena and Mouret, Jean-Baptiste},
  booktitle={2024 IEEE-RAS 23rd International Conference on Humanoid Robots (Humanoids)},
  pages={528--535},
  year={2024},
  organization={IEEE}
}

@inproceedings{li2025manidp,
  title={ManiDP: Manipulability-aware diffusion policy for posture-dependent bimanual manipulation},
  author={Li, Zhuo and Liu, Junjia and Li, Dianxi and Teng, Tao and Li, Miao and Calinon, Sylvain and Caldwell, Darwin and Chen, Fei},
  booktitle={2025 IEEE/RSJ International Conference on Intelligent Robots and Systems (IROS)},
  pages={9956--9962},
  year={2025},
  organization={IEEE}
}

@article{liu2026twindp3,
  title={{Twin-DP3}: View-Invariant {3D} Diffusion Policy With Digital Twin},
  author={Liu, Xinyu and Han, Feng and Cui, Jingzhi and Lou, Jiabin and Ding, Rong and Niu, Jianwei},
  journal={IEEE Robotics and Automation Letters},
  volume={11},
  number={3},
  year={2026},
  doi={10.1109/LRA.2026.3655308}
}

@article{xu2026hero,
  title={{HeRO}: Hierarchical {3D} Semantic Representation for Pose-aware Object Manipulation},
  author={Xu, Chongyang and Cheng, Shen and Li, Haipeng and Fan, Haoqiang and Feng, Ziliang and Liu, Shuaicheng},
  journal={arXiv preprint arXiv:2602.18817},
  year={2026}
}
\end{document}